\documentclass[11pt]{article}

\usepackage[final]{acl}

\usepackage{booktabs}
\usepackage{amsmath}
\usepackage{amssymb}
\usepackage{stfloats}
\usepackage{times}
\usepackage{latexsym}
\usepackage{hyperref}
\usepackage[T1]{fontenc}
\usepackage[utf8]{inputenc}
\usepackage{microtype}
\usepackage{inconsolata}
\usepackage{graphicx}
\usepackage{multirow}
\usepackage{tfrupee}
\usepackage{tabularx}

\usepackage{xcolor}

\definecolor{RiskCritical}{HTML}{8B0000}
\definecolor{RiskSevere}{HTML}{FF0000}   
\definecolor{RiskHigh}{HTML}{FF8C00}     
\definecolor{RiskModerate}{HTML}{FFD700} 
\definecolor{RiskLow}{HTML}{008000}      

\title{BetXplain: An Explanation-Annotated Dataset for Detecting Manipulative Betting Advertisements on Social Media}

\author{
MSVPJ Sathvik$^{1}$, Parmitha Vangapandu$^{2}$, Nishit Rane$^{2}$, Sathwik Narkedimilli$^{3}$,\\ \textbf{Mark Lee$^{1}$, and Akrati Saxena$^{4}$} \\
$^{1}$University of Birmingham, UK, 
$^{2}$IIIT Dharwad, India,
$^{3}$NUS, Singapore,\\
$^{4}$LIACS, Leiden University, The Netherlands \\
\texttt{msvpjsathvik@gmail.com,23bcs092@iiitdwd.ac.in,nishit.rane0201@gmail.com,} \\
\texttt{sathwik.narkedimilli@u.nus.edu,m.g.lee@bham.ac.uk,a.saxena@liacs.leidenuniv.nl} 
}

\begin{document}
\maketitle

\begin{abstract}
The promotion of betting applications on social media platforms has increased significantly in recent years. Many of these advertisements use persuasive techniques that may mislead users, encourage risky behavior, and potentially influence users’ mental well-being. However, research on the automated detection of manipulative and deceptive betting advertisements remains limited due to the lack of publicly available annotated datasets. In this work, we introduce a new dataset of betting-related advertisements collected from two widely used social media platforms, Instagram and Reddit. The advertisements were manually annotated for manipulative and deceptive advertising practices. In addition to classification labels, the dataset includes human-provided explanations that describe the reasoning behind each annotation, enabling research into explainable approaches to detecting manipulative advertising. Furthermore, we analyze the strategies commonly used in betting advertisements and examine how these persuasive tactics may impact users’ mental health. The proposed framework can also enable practical applications such as browser plugins that warn users about manipulative betting advertisements and automated web crawlers that help regulatory authorities monitor and detect such promotions online.
\end{abstract}

\section{Introduction}

Social media platforms have become major channels for digital advertising, with betting app promotions increasingly prevalent on Instagram and Reddit. These advertisements commonly employ persuasive messaging, promotional bonuses, and attractive marketing strategies to maximize user engagement. However, many betting ads incorporate manipulative or deceptive techniques that can mislead users and encourage risky financial behavior. The aggressive promotion of betting applications has raised significant societal concerns due to its association with financial losses, psychological distress, and gambling addiction. In Telangana alone, over 1000 suicides have reportedly been linked to betting apps\cite{IndianExpressBetting2025,BusinessStandardYouth2025}, highlighting the severe mental health risks associated with unrealistic portrayals of betting as an easy income source.

Certain betting platforms employ deceptive operational practices alongside persuasive advertising strategies. Some applications initially permit small withdrawals to build user trust but later restrict or disable withdrawals after larger balances accumulate, preventing users from accessing funds~\cite{avast2025bettingScams,forcepoint2026casinoScams}. Such practices constitute digital financial exploitation and raise serious consumer protection concerns. Regulatory challenges further intensify the issue, as betting is restricted or banned in India~\cite{PublicGamblingAct1867} and several Gulf countries~\cite{UAEFederalPenalCode1987,SaudiGamblingLaw,QatarPenalCode2004,KuwaitPenalCode1960} due to risks of financial harm and addiction. Nevertheless, betting advertisements continue to circulate across global social media platforms, bypassing regional restrictions and exposing users to manipulative promotions~\cite{ITAct2000}.

Research on automated detection of manipulative and deceptive betting advertisements remains limited due to the absence of publicly available social media datasets focused on betting promotions. Existing resources primarily analyze gambling behavior or betting statistics and rarely provide explanatory annotations for deceptive content. To address this gap, we introduce BetXplain\footnote{The dataset will be made publicly available upon acceptance. The access link has been omitted from this version to comply with the anonymized submission policy.} A dataset of Instagram and Reddit betting advertisements manually annotated as manipulative, deceptive, or responsible, with human-written explanations. We further analyze persuasion strategies, sentiment patterns, and mental health risks, and evaluate transformer-based and large language models for detecting manipulative advertisements.

The key contributions of the paper are as follows:

\begin{enumerate}
 \item This is the first work to develop methods for detecting manipulative betting apps.

 \item This paper is the first to propose a dataset on manipulative betting apps with explanations. 

 \item We have presented a novel analysis of how these betting apps impact mental health and other areas.
\end{enumerate}

\section{Related Work}

Online gambling and mobile betting platforms have rapidly expanded in recent years, transforming gambling from physical venues to continuously accessible digital environments. Mobile betting applications provide real-time wagering, in-play betting, and personalized promotional notifications that encourage frequent engagement~\cite{gainsbury2015online, abbott1991gambling}. This shift has been accompanied by aggressive marketing strategies that rely on targeted promotions, bonus incentives, and time-sensitive offers delivered through push notifications and advertisements~\cite{hing2014sports, lopez2018understanding}. Such strategies often frame gambling as entertaining and profitable while minimizing potential risks. The constant accessibility of mobile platforms further removes traditional barriers such as travel time and social oversight, increasing the likelihood of problematic gambling behavior~\cite{abbott2020changing}.

Gambling advertisements often employ persuasive strategies such as urgency cues (e.g., ``limited-time offers'' and ``boosted odds'') and reward framing that emphasize financial gains while minimizing losses~\cite{hing2014sports,newall2019dark,guerrero2018online,binde2014gambling}. By associating betting with sports fandom and entertainment, these advertisements may normalize gambling and increase participation among vulnerable populations~\cite{hing2014sports}. Problem gambling is associated with anxiety, depression, financial distress, and interpersonal conflict, while online environments may further amplify risks through persistent promotional exposure and distorted perceptions of winning probabilities~\cite{petry2005pathological,walker1996prevalence,gainsbury2015online,dowling2015prevalence, gorrepati2025mental}. Although NLP methods and transformer-based models, including BERT, RoBERTa, DistilBERT, ELECTRA, ALBERT, and GPT-4, have achieved strong results in harmful content detection~\cite{davidson2017automated,fortuna2018survey,schmidt2017survey,vidgen2021learning,devlin2019bert,liu2019roberta,sanh2019distilbert,clark2020electra,lan2019albert,brown2020language}, gambling advertisements remain underexplored, with no public dataset providing annotated explanations for manipulative betting promotions~\cite{hing2014sports,griffiths2019online}. This work addresses this gap through the proposed BetXplain dataset.

Gambling behavior has been widely studied in relation to addiction, cognitive bias, and reward-system dysregulation, with intermittent reward mechanisms shown to activate dopaminergic pathways similar to substance addiction, thereby reinforcing compulsive behavior~\cite{petry2005pathological,gainsbury2015online}. Gambling advertisements further exploit cognitive biases, including the illusion of control, loss aversion, and social proof, thereby increasing engagement and exposure risk, positioning betting advertisements as both linguistic artifacts and behavioral stimuli with measurable psychological impact. Related research on deceptive design identifies widespread use of dark patterns in online systems~\cite{mathur2019dark} and manipulative interface taxonomies~\cite{gray2018dark}. Studies on social media endorsements demonstrate how promotional content can obscure commercial intent~\cite{mathur2018endorsements}, while explanation-guided large language models improve the detection of misleading content~\cite{he2024harnessing}, motivating our explanation-annotated framework.

\section{Methodology}

This section describes the data collection process, annotation methodology, and dataset construction used in this study. It also outlines the procedures followed to ensure annotation quality and consistency.

% \begin{figure*}
% \centering
% \includegraphics[width=1\linewidth]{Betting_App_Methodology_Diagram.drawio.png}
% \caption{Archetecture Diagram}
% \label{fig:placeholder}
% \end{figure*}

\begin{figure*}
  \centering
  \setlength{\fboxsep}{0pt}
  \fbox{\includegraphics[width=\dimexpr 0.72\linewidth - 2\fboxrule \relax]{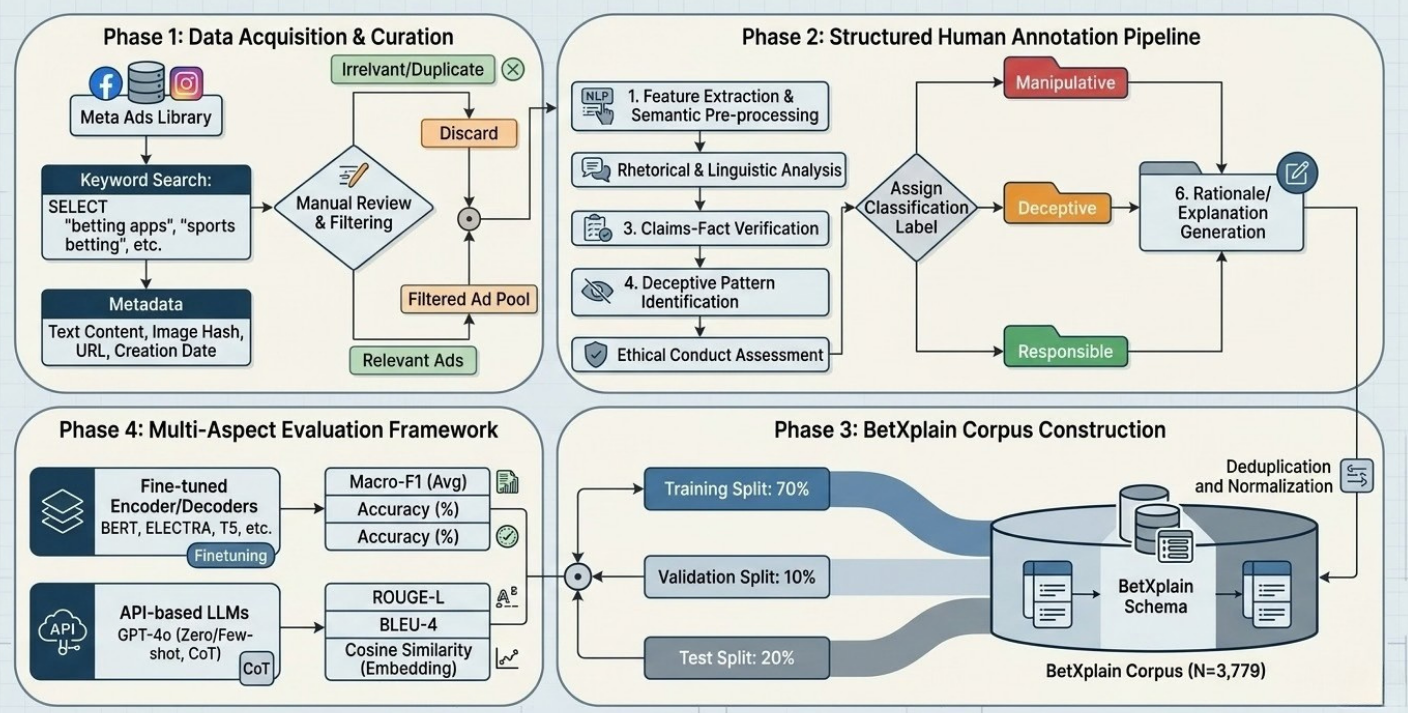}}
  \caption{Overview of the BetXplain framework, illustrating the pipeline from betting advertisement data collection and manual annotation to corpus construction and multi-aspect evaluation using transformer models and LLM-based explanation metrics.}
  \label{fig:placeholder}
\end{figure*}

\subsection{Data Collection}
To construct the dataset, we collected betting-related advertisements from social media platforms using the Meta Ads Library, which provides access to publicly available advertisements running across Meta platforms. The data collection focused on advertisements for betting applications and related services. To identify relevant advertisements, we queried the platform using keywords such as ``betting apps”, ``online betting”, ``sports betting”, and related terms. These queries returned advertisements associated with betting platforms that were actively promoted on social media.

For each advertisement retrieved through the Meta Ads Library, we collected the advertisement link and associated metadata. The advertisements were then manually reviewed to verify that they were related to betting applications or betting services. Non-relevant advertisements or duplicate entries were removed during this filtering stage to ensure that the dataset contained only relevant betting-related promotional content. After collecting the advertisements, they were assigned to two volunteer data annotators who contributed to this project and are also co-authors of this paper, with no financial compensation provided. The annotators manually reviewed each instance and labeled it according to the advertising strategy employed, categorizing each advertisement as manipulative, deceptive, or responsible.

\begin{enumerate}

\item Manipulative advertisements refer to promotional content that uses persuasive or psychological tactics to influence user behavior. These advertisements often exaggerate potential rewards, create a sense of urgency, or portray betting as an easy way to earn money. Such strategies may pressure users into participating in betting activities without adequately presenting the associated risks.

\item Deceptive advertisements refer to advertisements that present misleading or false information about betting platforms. These may include unrealistic claims about guaranteed winnings, misleading promotional offers, or advertisements linked to platforms that engage in questionable practices, such as restricting withdrawals after users deposit funds.

\item Responsible advertisements are betting promotions that present information in a neutral or transparent manner. These advertisements typically avoid exaggerated claims and may include responsible messaging or warnings regarding gambling behavior.
\end{enumerate}

To ensure the dataset's reliability, all advertisements were manually verified during annotation. In addition to assigning category labels, annotators also provided textual explanations describing the reasoning behind each annotation. These explanations highlight the specific elements within an advertisement that indicate manipulative or deceptive behavior, such as exaggerated reward framing, urgency-driven messaging, or misleading promotional claims. The inclusion of explanation annotations enhances the dataset's interpretability and enables research on explainable detection models for manipulative advertising. By capturing both classification labels and the reasoning behind them, the dataset supports the development of automated systems that can not only detect harmful betting advertisements but also explain why they are considered manipulative or deceptive.

\subsection{Data Annotation Guidelines}
To maintain consistent labeling across the dataset, annotators followed the following guidelines:

\begin{enumerate}

\item Content Verification: 
Annotators first verified whether the advertisement was genuinely related to betting applications or betting services. Non-relevant advertisements were excluded from the dataset.

\item Identification of Persuasive Language:
Annotators examined whether the advertisement used persuasive or emotional language designed to strongly influence user behavior.

\item Detection of Misleading Claims:
Annotators checked for misleading statements such as guaranteed profits, unrealistic rewards, or exaggerated success rates.

\item Evaluation of Promotional Tactics:
Advertisements were examined for tactics such as urgency-based promotions, exaggerated bonuses, or claims that portray betting as an easy source of income.

\item Assessment of Platform Practices:
When possible, annotators reviewed linked platforms to determine whether the advertisement promoted services associated with questionable practices, such as restricted withdrawals.

\item Explanation Annotation:
For each labeled advertisement, annotators provided a short explanation describing why the advertisement was categorized as manipulative, deceptive, or responsible. These explanations highlight the key features that influenced the labeling decision.

\end{enumerate}

The use of multiple annotators and structured annotation guidelines helped ensure the dataset's reliability and quality. The inclusion of explanation annotations also enables research on explainable detection models, allowing automated systems to identify manipulative betting advertisements while providing clear reasoning behind their predictions.

\subsection{Dataset Overview}
The proposed dataset, \textit{BetXplain}, contains betting-related advertisements collected from public social media platforms, primarily Instagram, to study manipulative and deceptive promotional strategies used in betting applications. Each instance in the dataset includes five attributes: \textit{text}, \textit{link}, \textit{category}, \textit{explanation}, and \textit{promotion label} as shown in Table~\ref{tab:overview}. The \textit{text} field represents the advertisement caption or promotional message, while the \textit{link} field contains the link to the ad. The \textit{category} attribute identifies the type of betting activity being promoted, such as gambling or colour prediction games. The \textit{promotion label} classifies the advertisement as \textit{manipulative}, \textit{deceptive}, or \textit{responsible}. In addition to these labels, the dataset includes human-written \textit{explanations} describing the reasoning behind each annotation, highlighting persuasive language, exaggerated reward framing, or the absence of risk disclosure. This structured annotation enables the development of automated models that detect harmful betting advertisements while also supporting explainable analysis of promotional tactics.

\subsection{Data Statistics}

The BetXplain dataset contains 3,779 advertisements after removing 216 duplicate entries from the original 4,000 collected. The label distribution is moderately imbalanced, with responsible promotion forming the largest class (49.6\%), followed by manipulative promotion (39.9\%), and deceptive promotion as the minority class (10.5\%). This distribution reflects real-world advertising patterns, where explicit false claims are less common than psychologically manipulative but technically permissible promotions. Advertisement texts are generally short, with a median length of 13 words and a mean of 30.1 words ($\sigma = 39.2$), consistent with the concise nature of social media promotional captions. The dataset spans 19 advertising categories, including gambling-related content such as slots, poker, roulette, color prediction, and casino, as well as non-gambling sports content like cricket, football, tennis, and basketball used as responsible promotion baselines.

\begin{table*}[t]
\centering
\scriptsize
\resizebox{0.8\textwidth}{!}{%
\begin{tabular}{p{3.2cm}|p{2.5cm}|p{1.8cm}|p{1.8cm}|p{4.5cm}}
\toprule
\textbf{Text} & \textbf{Link} & \textbf{Category} & \textbf{Label} & \textbf{Explanation} \\
\midrule

Join India's Best Betting Platform With Self Deposit/Self Withdrawal Facility, Welcome Bonus: 99\rupee, Min Deposit: 100\rupee, Every Deposit Bonus: 5\%, Referral Bonus: 5\%, 24x7 Withdrawal &
\url{https://www.instagram.com/reels/DKrX-UsS-Jd/\%7D} &
casino &
manipulative &
Uses exclusivity and status-based language to create aspirational appeal without clarifying requirements, costs, or risks. Emotional and status-driven persuasion without overt deception. \\
\midrule

Vote below fellas. Let's see it &
\url{https://www.instagram.com/reels/DFJHp4xTo1l/} &
hockey (non-gambling) &
responsible &
Purely informational sports content with no exaggerated claims, manipulative tactics, or misleading financial statements. \\
\midrule

Online earning game mines. Online job. Real money. Winning game. Aviator online game. Betting game. &
\url{https://www.instagram.com/reels/DHIVml2ISSn/} &
mines &
manipulative &
The post frames gambling as a legitimate income source using terms like `online job' and `real money' without any risk disclosure or responsible gambling messaging, exploiting financial aspirations through emotionally charged language and symbols. \\
\bottomrule

\end{tabular}%
}
\caption{Sample instances from the BetXplain dataset illustrating each promotion label.}
\label{tab:overview}
\end{table*}

\section{Experimental Results}
In this section, we present the experimental setup and evaluate the performance of different models on the proposed dataset. We also analyze the results using standard classification and explanation quality metrics.

\subsection{Experimental Setup}

\paragraph{Dataset and Splits.}
Our dataset comprises 3{,}779 social media posts annotated across three classes: \textit{manipulative promotion} (1{,}507 samples, 39.9\%), \textit{deceptive promotion} (396 samples, 10.5\%), and \textit{responsible promotion} (1{,}876 samples, 49.6\%). After removing 216 duplicate entries and verifying zero train-test overlap, we partition the data into training (70\%, 2{,}645 samples), validation (10\%, 378 samples), and test (20\%, 756 samples) splits using stratified sampling to preserve class proportions.
 
\paragraph{Models.}
We evaluate two broad categories of models: fine-tuned transformer models and closed-source API models. For \textit{fine-tuned transformer models}, we consider encoder-only models—BERT \cite{devlin2019bert}, RoBERTa \cite{liu2019roberta}, DistilBERT \cite{sanh2019distilbert}, ELECTRA \cite{clark2020electra}, and ALBERT \cite{lan2019albert}-a long-context encoder, Longformer
\cite{beltagy2020longformer} and sequence-to-sequence models BART \cite{lewis2020bart} and T5 \cite{raffel2020exploring}. For \textit{closed-source API models}, we evaluate GPT-4o under three prompting strategies: zero-shot, few-shot (5 examples), and chain-of-thought (CoT).
 
\paragraph{Training Details.}
All fine-tuned models are trained for 5 epochs using the AdamW optimizer \cite{loshchilov2017decoupled} with weight decay 0.01 and a linear warmup schedule covering 10\% of total steps with a learning rate of $2 \times 10^{-5}$. Automatic Mixed Precision (AMP) is enabled for all models except T5, which requires full float32 precision. To address class imbalance, particularly the underrepresented deceptive class ($\approx$10\%), we apply inverse-frequency class weighting to the cross-entropy loss. All models use a maximum sequence length of 256 tokens and a batch size of 16. The best checkpoint is selected based on macro-F1 on the validation set. Experiments are conducted on an NVIDIA GeForce RTX 5060 Ti (16 GB). We report accuracy, macro-F1, weighted-F1, and micro-F1 on the held-out test set. Macro-F1 is our primary metric given the class imbalance.
 
\subsection{Classification Results}
 
Table~\ref{tab:main_results} reports classification performance across all models.
 
\begin{table}[t]
\centering
\scriptsize
\setlength{\tabcolsep}{3pt}
\resizebox{\columnwidth}{!}{
\begin{tabular}{llcccc}
\toprule
\textbf{Group} & \textbf{Model} & \textbf{Acc} & \textbf{Ma-F1} & \textbf{W-F1} & \textbf{Mi-F1} \\
\midrule
\multirow{5}{*}{Encoder}
 & BERT & 0.8148 & 0.6752 & 0.8171 & 0.8148 \\
 & RoBERTa & 0.7947 & 0.6685 & 0.8057 & 0.7947 \\
 & DistilBERT & 0.8223 & 0.6899 & 0.8257 & 0.8223 \\
 & ELECTRA & \textbf{0.8335} & \textbf{0.6946} & \textbf{0.8326} & \textbf{0.8335} \\
 & ALBERT & 0.8323 & 0.6876 & 0.8286 & 0.8323 \\
\midrule
Long-context
 & Longformer & 0.8511 & 0.6711 & 0.8351 & 0.8511 \\
\midrule
\multirow{2}{*}{Enc-Dec}
 & BART & 0.8010 & 0.6763 & 0.8095 & 0.8010 \\
 & T5 & 0.8223 & 0.6519 & 0.8126 & 0.8223 \\
\midrule
\multirow{3}{*}{API}
 & GPT-4o (zero-shot) & 0.8123 & 0.6871 & 0.8098 & 0.8123 \\
 & GPT-4o (few-shot) & \underline{0.8210} & \underline{0.6898} & 0.8151 & \underline{0.8210} \\
 & GPT-4o (CoT) & 0.8181 & 0.6870 & \underline{0.8158} & 0.8181 \\
\bottomrule
\end{tabular}
}
\caption{Classification results on the test set (756 samples). \textbf{Bold}: best fine-tuned model per metric. All metrics computed with 95\% bootstrap confidence intervals (CI width $\approx$0.05--0.08); see supplementary materials for full CIs.}
\label{tab:main_results}
\end{table}
 
\paragraph{Fine-tuned Models.}
Among fine-tuned models, ELECTRA achieves the highest macro-F1 (0.6946), followed by DistilBERT (0.6899) and ALBERT (0.6876). ELECTRA’s discriminator-based pretraining, which classifies real versus replaced tokens across all positions, appears to produce more discriminative representations for fine-grained promotion detection. Fine-tuned models span a macro-F1 range of 0.651–0.695, with T5 performing weakest (0.6519) among encoder-decoder architectures. Longformer achieves the highest accuracy (0.8511) and macro-F1 (0.6711), indicating benefits from global attention beyond long-context modeling. Although the median text length is 13 words, 5.5\% of samples exceed 128 tokens, and Appendix~\ref{app:longformer} shows gains arise from global attention rather than increased context length.
 
\paragraph{API Models.}
GPT-4o with few-shot prompting achieves the highest accuracy among API configurations
(0.8210), while chain-of-thought achieves the highest macro-F1 (0.6870), yet this remains below the best fine-tuned model (ELECTRA: 0.6946). The structured reasoning of CoT better handles the underrepresented \textit{deceptive promotion} class than few-shot methods, which tend to steer the model toward majority-class predictions. Across all three GPT-4o strategies, the gap between weighted-F1 ($\approx$0.81) and macro-F1 ($\approx$0.69) is notably larger than for fine-tuned models, confirming that GPT-4o systematically under-identifies the \textit{deceptive promotion} class. This underscores the value of fine-tuning on domain-specific annotated data, even against a capable frontier model.

\subsection{Open-Source LLM Evaluation}

To evaluate accessibility beyond proprietary models, we tested two open-source instruction-tuned LLMs, LLaMA-3-8B-Instruct and Mistral-7B-Instruct-v0.3, under zero-shot and few-shot (5 examples) prompting. Table~\ref{tab:opensource} shows that both models perform 3--7\% below GPT-4o, consistent with the known capability gap between frontier and mid-size models. Few-shot prompting improves performance by ~2-3\%, confirming that in-context learning is effective even for smaller models. However, even the best open-source result (LLaMA-3 few-shot, macro-F1 = 0.6798) remains below fine-tuned ELECTRA (0.6946), reinforcing the value of domain-specific fine-tuning.

\begin{table}[t]
\centering
\footnotesize
\setlength{\tabcolsep}{3.5pt}
\begin{tabular}{lcccc}
\toprule
\textbf{Model} & \textbf{Acc} & \textbf{Ma} & \textbf{W} & \textbf{Mi} \\
\midrule
\multicolumn{5}{l}{\textit{GPT-4o}} \\
Zero-Shot  & 0.812 & 0.687 & 0.810 & 0.812 \\
Few-Shot  & 0.821 & 0.690 & 0.815 & 0.821 \\
CoT & 0.818 & 0.687 & 0.816 & 0.818 \\
\midrule
\multicolumn{5}{l}{\textit{Open LLMs}} \\
LLaMA-3-8B (zero-shot) & 0.747 & 0.638 & 0.743 & 0.747 \\
LLaMA-3-8B (few-shot) & 0.779 & 0.680 & 0.776 & 0.779 \\
Mistral-7B (zero-shot)   & 0.729 & 0.616 & 0.722 & 0.729 \\
Mistral-7B (few-shot)   & 0.767 & 0.665 & 0.763 & 0.767 \\
\bottomrule
\end{tabular}
\caption{Open-source LLM performance compared with GPT-4o. ZS: zero-shot, FS: few-shot.}
\label{tab:opensource}
\end{table}
 
\subsection{Explanation Quality}
 
Table~\ref{tab:explanation} reports explanation quality metrics comparing GPT-4o-generated rationales against human-authored gold explanations.
 
\begin{table}[t]
\centering
\small
\setlength{\tabcolsep}{5pt}
\begin{tabular}{lccccc}
\toprule
\textbf{Model} & \textbf{R-1} & \textbf{R-L} & \textbf{BLEU-1} & \textbf{Cosine} \\
\midrule
GPT-4o (zero-shot) & 0.2289 & 0.1852 & 0.0127 & 0.5398 \\
GPT-4o (few-shot)  & 0.1528 & 0.1144 & 0.0037 & 0.4964 \\
GPT-4o (CoT)   & \textbf{0.2707} & \textbf{0.2144} & \textbf{0.0173} & \textbf{0.5776} \\
\bottomrule
\end{tabular}
\caption{Explanation quality metrics against gold human rationales.
R-1/R-L: ROUGE-1/ROUGE-L F1. BLEU-1: unigram BLEU.
Cosine: sentence embedding cosine similarity (all-mpnet-base-v2).}
\label{tab:explanation}
\end{table}
 
CoT prompting yields the most semantically aligned explanations, achieving the highest ROUGE-1 (0.2707), ROUGE-L (0.2144), and cosine similarity (0.5776) scores. The comparatively high cosine similarity values (0.50--0.58) indicate strong semantic alignment between generated explanations and gold annotations despite lexical variation. In contrast, few-shot prompting produces the weakest explanations despite achieving the highest classification accuracy, suggesting that example conditioning biases the model toward replicating classification behavior rather than generating diverse, well-reasoned rationales. Exact match remains zero across all configurations, confirming that the generated explanations are stylistically distinct from the concise and structured gold rationales.

\subsection{Linking Classification Performance to Persuasion Patterns}

To connect the classification results (Section 4) with the linguistic analysis (Section 5), we analyze per-class F1 performance. Table~\ref{tab:perclass} shows that the deceptive class achieves the lowest F1 across models (0.43--0.51), consistent with Section 5.2, where deceptive ads share 72\% of their financial coercion vocabulary with manipulative ads, making discrimination inherently difficult. The dominant manipulative $\rightarrow$ responsible for confusion aligns with Section 5.1, which identifies play'' as a common cross-category leisure-framing term. ELECTRA’s discriminator pretraining captures subtle token-level distinctions, such as guaranteed'' versus ``boosted,'' explaining its +1.9\% macro-F1 improvement over BERT.

\begin{table}[t]
\centering
\footnotesize
\setlength{\tabcolsep}{4pt}
\begin{tabular}{lcccc}
\toprule
\textbf{Model} & \textbf{Manip.} & \textbf{Decept.} & \textbf{Resp.} & \textbf{Ma-F1} \\
\midrule
ELECTRA      & 0.828 & 0.511 & 0.879 & 0.739 \\
BERT         & 0.802 & 0.428 & 0.866 & 0.699 \\
GPT-4o (CoT) & 0.813 & 0.469 & 0.865 & 0.716 \\
\bottomrule
\end{tabular}
\caption{Per-class F1 breakdown for selected models.}
\label{tab:perclass}
\end{table}

 \begin{figure*}[t]
  \centering
  \includegraphics[width=0.9\linewidth]{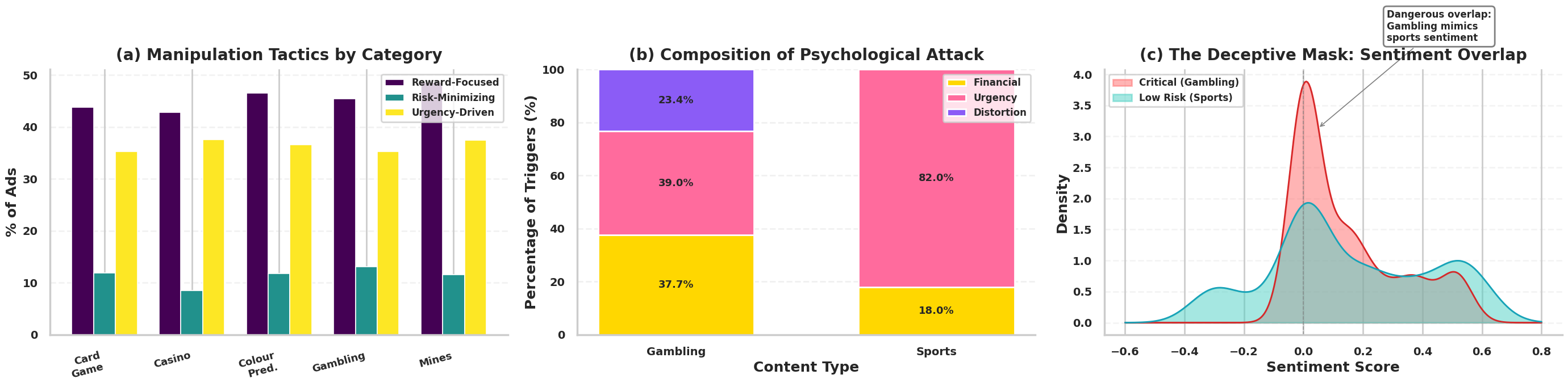}
  \caption{Psychological manipulation, architecture, and mental health impact(a) Keyword frequency heatmap across betting 
  categories. (b) Psychological trigger composition in gambling 
  versus non-gambling content. (c) Sentiment KDE overlap confirming 
  the Deceptive Positivity effect.}
  \label{fig:rq1}
\end{figure*}

\begin{figure*}[t]
  \centering
  \includegraphics[width=0.85\linewidth]{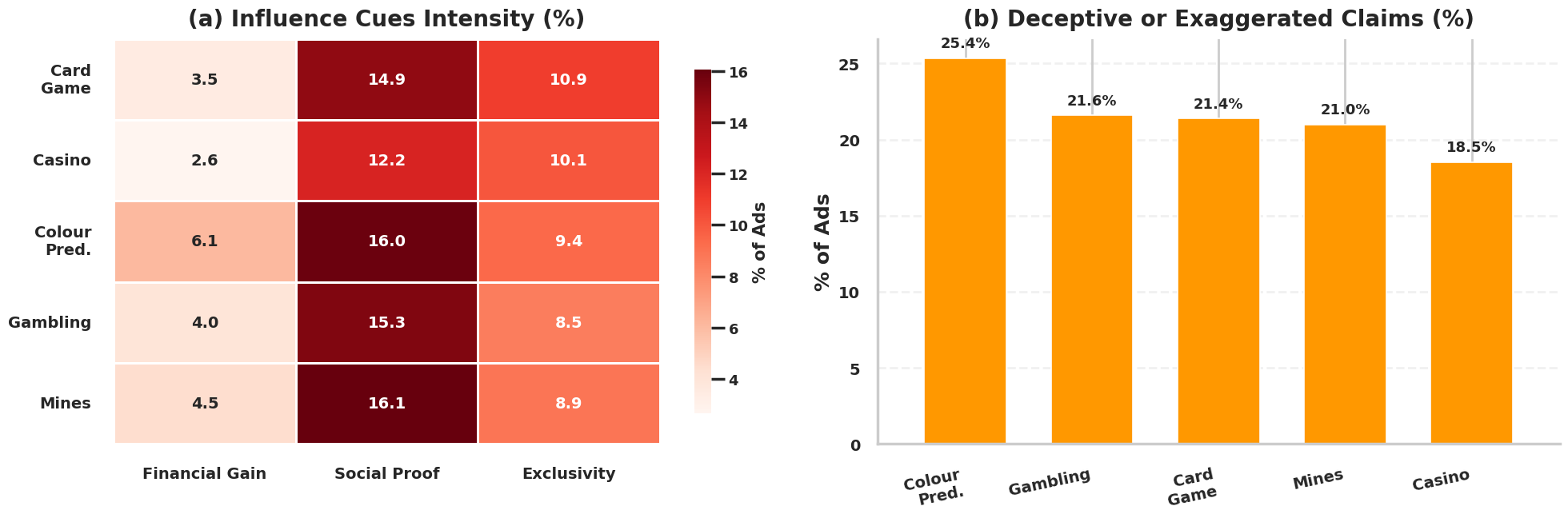}
  \caption{Influence cues and deceptive claims across categories: (a) Influence cues intensity by category. (b) 
  Prevalence of deceptive claims by category, with color prediction 
  highest at 25.4\%.}
  \label{fig:rq3}
\end{figure*}
\begin{figure*}[t]
  \centering
  \includegraphics[width=0.85\linewidth]{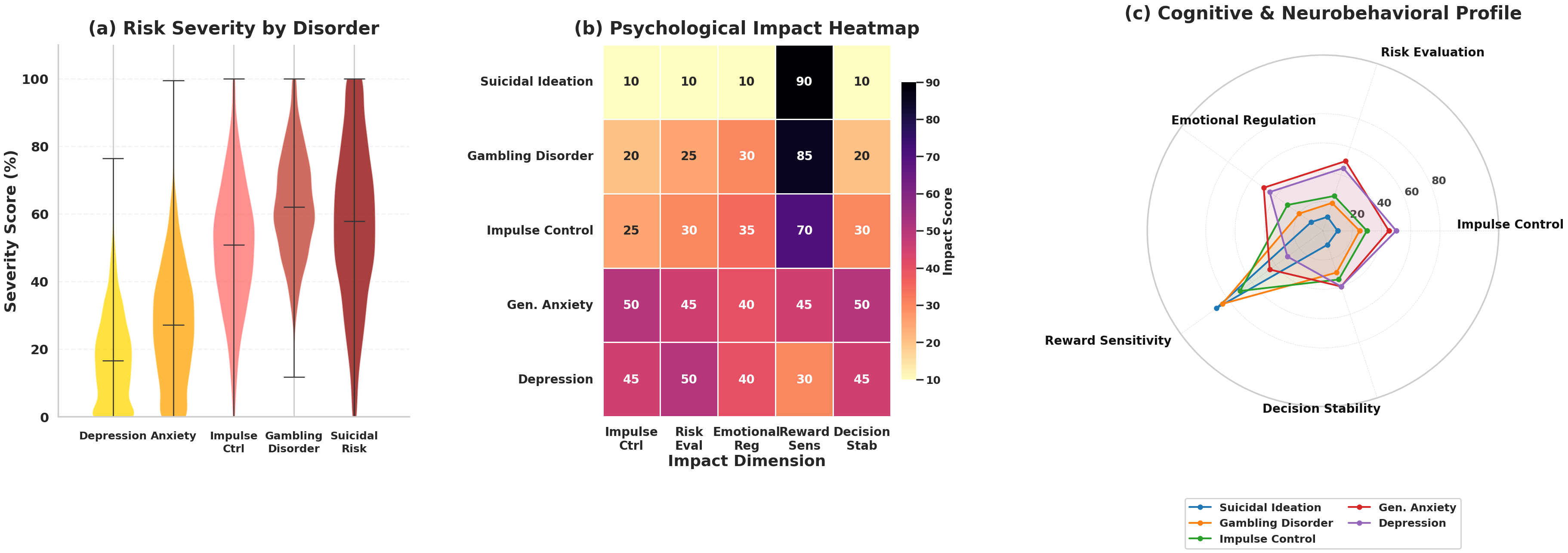}
  \caption{Clinical risk profile: (a) Risk severity distributions by disorder. 
  (b) Psychological impact heatmap across five cognitive dimensions. 
  (c) Neurobehavioral radar chart showing the asymmetric profile of 
  gambling disorder and suicidal ideation risk.}
  \label{fig:rq4}
\end{figure*}

\section{Discussion}
In this section, we interpret the experimental findings and examine the broader implications of the results. We analyze the persuasion patterns, emotional framing, and psychological triggers present in betting advertisements and discuss how these elements contribute to manipulative promotional strategies.

\subsection{Category-Wise Persuasion Strategies}

As shown in Figure~\ref{fig:rq1}(a), we analyze the distribution of manipulation tactics across betting advertisement categories. The results indicate that \textit{reward-focused messaging} dominates across all categories, accounting for approximately 42–47\% of advertisements. These messages emphasize potential gains, bonuses, and winning opportunities, reinforcing aspirational appeal.

\textit{Urgency-driven tactics} form the second most common strategy (approximately 35–38\%), using cues such as limited-time offers and immediate rewards to accelerate decision-making and reduce deliberation.

In contrast, \textit{risk-minimizing language} is consistently underrepresented (8–13\%), indicating that advertisements systematically downplay or omit potential losses and financial risks. This imbalance suggests that betting promotions are strategically optimized to amplify perceived rewards while suppressing risk awareness across all categories.

Figure~\ref{fig:rq1}(b) shows that gambling content deploys financial coercion ($\sim$56\%), urgency ($\sim$26\%), and cognitive distortion ($\sim$18\%), a combination absent in sports content entirely. This asymmetry reflects a shared operator template rather than an independent 
optimization \cite{petry2005pathological}, and validates training a single unified model on BetXplain. It also explains the macro-F1 range among fine-tuned models (0.651-0.695): the shared persuasion vocabulary is detectable, but subtler deceptive signals remain difficult to resolve.

\subsection{Influence Cues and Deceptive Claims}

Social proof dominates across all categories (12-16\%), 
Figure~\ref{fig:rq3}(a), while explicit financial claims remain rare (2.6-6.1\%). This is not a coincidence; social proof implies community participation without falsifiable claims, making it harder to flag while achieving the same legitimizing effect. Deceptive claims appear in 18.5-25.4\% of ads (Figure~\ref{fig:rq3}(b)), with color prediction highest at 25.4\%, consistent with its strategy of misrepresenting a chance game as skill-tractable. Of 401 financial 
claim ads, nearly half used urgency-to-deposit framing risk is not disclosed; it is crowded out \cite{dowling2015prevalence}.

\subsection{Clinical Risk Severity and Mental Health Impact}

To construct the clinical risk profile, we map linguistic and psychological features extracted from advertisements to established behavioral risk factors reported in psychological and psychiatric literature on gambling addiction. Specifically, each advertisement is analyzed for the presence of key trigger categories, including financial coercion, urgency, reward framing, and cognitive distortion. These triggers are then associated with clinically recognized mental health outcomes such as gambling disorder, anxiety, depression, and suicidal ideation based on prior empirical studies~\cite{petry2005pathological, gainsbury2015online}. Risk severity scores are computed by aggregating the intensity and co-occurrence of these triggers, producing a normalized severity scale ranging from low to critical risk. This mapping allows us to interpret advertising strategies not only as linguistic patterns but as potential contributors to psychological harm.

Figure~\ref{fig:rq4}(a) reveals a clear severity escalation gradient, with gambling disorder and suicidal ideation clustering at the severe end (9.0--9.5/10). The impact heatmap in Figure~\ref{fig:rq4}(b) identifies reward sensitivity dysregulation as the dominant factor, with suicidal ideation and gambling disorder scoring 90 and 85, respectively, while showing regulatory collapse across other dimensions. Figure~\ref{fig:rq4}(c) further confirms a distinct asymmetric neurobehavioral profile relative to anxiety and depression \cite{petry2005pathological}. Among 207 family-related ads, nearly equal proportions employed aspirational and threat-based framing, intensifying emotional distress during losses. These findings position manipulative betting advertisement detection as both a linguistic and a mental health intervention task, with BetXplain providing transparent, auditable reasoning annotations.

\subsection{Real World Applications}

The proposed framework enables multiple real-world applications for mitigating manipulative betting advertisements on digital platforms. It can be integrated into browser or mobile plugins to automatically analyze betting advertisements on social media and warn users about manipulative or deceptive content by highlighting tactics such as exaggerated reward claims and urgency-driven messaging. The framework can also support automated web crawlers for governments and regulatory agencies to monitor betting promotions and identify suspicious advertisements at scale. Additionally, it can enhance social media moderation systems and advertising compliance tools by automatically flagging potentially misleading betting promotions before they are widely disseminated.

\section{Conclusion}
In this work, we introduced \textbf{BetXplain}, an explanation-annotated dataset designed to support the detection of manipulative and deceptive betting advertisements on social media. The dataset provides manually verified advertisements labeled as manipulative, deceptive, or responsible, along with human-authored explanations that capture the reasoning behind each annotation. Through experiments with transformer-based models and large-language-model prompting strategies, we demonstrate that domain-specific training enables effective identification of persuasive and misleading betting promotions. Our analysis further highlights common persuasive strategies used in betting advertisements, including exaggerated reward claims and urgency-driven messaging. Overall, BetXplain provides a new resource for studying harmful advertising practices and supports future research on explainable detection systems to promote safer and more responsible online advertising environments.

\section*{Limitations}
The scope of this study focuses on betting advertisements collected from selected social media platforms where such promotions are commonly observed. While these platforms are important channels for digital advertising, betting promotions may also appear in other online environments and regional ecosystems. Expanding the dataset to include advertisements from a broader range of platforms and geographic contexts could further enrich the diversity of promotional strategies captured and support the development of more globally robust detection systems. In addition, the current dataset focuses primarily on English-language advertisements. Betting promotions often vary across languages and cultural contexts, and extending the dataset to multilingual settings could enable the study of how persuasive strategies differ across regions and regulatory environments.

Finally, BetXplain provides concise, human-authored explanations that highlight the reasoning behind each annotation. These explanations support research on explainable detection models while maintaining annotation consistency. Future work may explore richer explanation formats or additional annotation perspectives that capture more detailed reasoning processes. These directions represent natural extensions of the present work, highlight opportunities to further advance research on responsible digital advertising and explainable harmful-content detection, and explainable harmful-content detection.

\section*{Ethics Statement}
We, the authors of this work, affirm that the research presented in this paper was conducted in accordance with ethical standards and responsible research practices. Throughout this study, we ensured that the collection and analysis of betting advertisements were conducted using publicly accessible sources and strictly for academic research purposes to understand and detect manipulative betting promotions on social media platforms. We acknowledge that certain limitations may arise from manual annotation processes and from the stochastic behavior inherent in modern machine learning and large language models. To address these concerns, we followed structured annotation guidelines, involved multiple annotators to maintain consistency, and documented the dataset construction, annotation procedures, model configurations, and experimental settings to support transparency and reproducibility of our work. We also maintained controlled experimental settings during model evaluation to minimize variability in results. The dataset contains betting-related promotional content that may include persuasive language, which could be sensitive for vulnerable individuals; however, it is used strictly for research, and we explicitly oppose and do not promote such activities. Annotation subjectivity and data-driven biases may affect model generalization across contexts. Additionally, links to public advertisements require responsible handling, and automated systems derived from this work should be used with appropriate human oversight and ethical safeguards. For coding assistance, AI-based coding tools were utilized; however, the conceptualization of the research problem, dataset creation, methodological design, experimentation, and analysis were conducted independently by the authors.

\bibliography{ref}

\clearpage

\appendix

\section*{Appendix}
 
\vspace{1em}

% ============================================================
% Appendix: Inter-Annotator Agreement and Annotation Reliability
% ============================================================

\section{Inter-Annotator Agreement Scores}
\label{app:iaa}

The annotation process was conducted by a team of four annotators who independently reviewed betting-related advertisements collected from public social media platforms. Since the distinction between manipulative, deceptive, and responsible promotional strategies can be subjective, detailed annotation guidelines were established before annotation to improve consistency among annotators.

Each annotator participated in an initial pilot annotation phase involving a subset of advertisements. Disagreements were discussed collaboratively to refine category boundaries and align annotation standards before the full annotation process began.

Advertisements were labeled according to the following promotion categories:

\begin{itemize}

\item \textbf{Manipulative Promotion (MP):}
Advertisements using emotionally persuasive or psychologically coercive tactics such as urgency framing, exaggerated rewards, aspirational language, fear of missing out (FOMO), or financial pressure are intended to influence user behavior.

\item \textbf{Deceptive Promotion (DP):}
Advertisements containing misleading or potentially false claims, including unrealistic guarantees of profit, exaggerated winning probabilities, deceptive promotional offers, or misleading financial representations.

\item \textbf{Responsible Promotion (RP):}
Advertisements presenting betting-related information in a comparatively transparent or neutral manner without strong manipulative framing or deceptive claims.

\end{itemize}

In addition to promotion labels, advertisements were grouped into 5 betting-related categories: casino promotions, sports betting, color prediction games, poker/card games, and miscellaneous betting applications.

To evaluate annotation reliability, we computed multiple inter-annotator agreement (IAA) metrics, including Cohen's $\kappa$, Fleiss' $\kappa$, and Krippendorff's $\alpha$. Pairwise Cohen's $\kappa$ values were computed between annotator pairs, while Fleiss' $\kappa$ and Krippendorff's $\alpha$ were computed across all annotators collectively.

\begin{table}[h!]
\centering
\small
\setlength{\tabcolsep}{4pt}
\renewcommand{\arraystretch}{1.2}
\resizebox{\linewidth}{!}{
\begin{tabular}{lccc}
\toprule
\textbf{Annotator Pair} &
\textbf{Krippendorff's $\alpha$} &
\textbf{Cohen's $\kappa$} &
\textbf{Fleiss' $\kappa$} \\
\midrule

(1,2) & 0.8127 & 0.8264 & --- \\
(1,3) & 0.7841 & 0.7989 & --- \\
(1,4) & 0.7365 & 0.7498 & --- \\
(2,3) & 0.7694 & 0.7816 & --- \\
(2,4) & 0.7218 & 0.7352 & --- \\
(3,4) & 0.7953 & 0.8087 & --- \\

\midrule

(1,2,3,4) & 0.7700 & --- & 0.7526 \\

\bottomrule
\end{tabular}
}
\caption{Inter-Annotator Agreement Scores}
\label{tab:iaa_appendix}
\end{table}

The agreement scores indicate moderate-to-strong consistency across annotator pairs. The averaged pairwise Cohen's $\kappa$ score was 0.7834, while the collective Krippendorff's $\alpha$ and Fleiss' $\kappa$ scores were 0.7700 and 0.7526, respectively. These values are consistent with complex annotation tasks that involve subjective distinctions among manipulative, deceptive, and responsible promotional strategies.

The comparatively lower agreement observed for some annotator pairs reflects the inherent ambiguity involved in distinguishing emotionally persuasive advertising language from explicitly deceptive claims. Advertisements involving aspirational financial messaging, urgency-based promotions, and implied earning opportunities produced the highest disagreement rates. All annotation disagreements were resolved through collaborative discussion and consensus review prior to finalizing the dataset.

\section{Cross-Validation Robustness}

To verify that the single-split results are not artifacts of a favorable partition, we performed 5-fold stratified cross-validation across all eight fine-tuned models. Table~\ref{tab:kfold} reports mean performance and standard deviation across folds. The model ranking is preserved: ELECTRA remains the top model by macro-F1, with standard deviations ranging from 0.007 to 0.014, indicating stable generalization despite the moderate dataset size. Cross-validation means are within 1--2\% of the single-split results, confirming that the test partition is representative.

\begin{table}[t]
\centering
\footnotesize
\setlength{\tabcolsep}{3pt}
\begin{tabular}{lcccc}
\toprule
\textbf{Model} & \textbf{Acc} & \textbf{Ma} & \textbf{W} & \textbf{Rep.} \\
\midrule
BERT    & 0.784$\pm$.008 & 0.688$\pm$.008 & 0.780$\pm$.008 & 0.675 \\
RoBERTa & 0.774$\pm$.011 & 0.667$\pm$.014 & 0.768$\pm$.011 & 0.669 \\
DistilB & 0.826$\pm$.010 & 0.736$\pm$.009 & 0.820$\pm$.010 & 0.690 \\
ELECTRA & \textbf{0.829$\pm$.009} & \textbf{0.739$\pm$.010} & \textbf{0.824$\pm$.010} & \textbf{0.695} \\
ALBERT  & 0.824$\pm$.012 & 0.728$\pm$.011 & 0.818$\pm$.011 & 0.688 \\
Longf.  & 0.835$\pm$.012 & 0.726$\pm$.011 & 0.825$\pm$.011 & 0.671 \\
BART    & 0.787$\pm$.007 & 0.691$\pm$.008 & 0.783$\pm$.007 & 0.676 \\
T5      & 0.812$\pm$.007 & 0.694$\pm$.011 & 0.803$\pm$.007 & 0.652 \\
\bottomrule
\end{tabular}
\caption{5-fold cross-validation results (mean$\pm$std) and reported macro-F1 scores. Acc: Accuracy, Ma: Macro-F1, W: Weighted-F1, Rep.: Reported single-split Macro-F1, Longf.: Longformer, DistilB: DistilBERT.}
\label{tab:kfold}
\end{table}

We conducted McNemar's test and paired bootstrap tests (10,000 resamples) with Benjamini-Hochberg correction, comparing all models against ELECTRA. After correction for multiple comparisons, no pairwise difference reaches statistical significance at $\alpha = 0.05$. This is expected given the moderate test set size (756 samples) and the narrow macro-F1 range (0.651-0.695). The models form a competitive cluster, and deployment decisions should consider computational constraints alongside raw metrics.

\section{LLM-as-a-Judge Consistency Analysis}
\label{app:llmjudge}

To further evaluate the consistency of the annotation framework, we additionally explored an LLM-as-a-Judge setting using GPT-4o. A randomly selected subset of advertisements was independently evaluated by the model, using the same annotation guidelines as those provided to human annotators.

The LLM-generated labels were compared against the finalized consensus annotations created by the human annotators. Agreement analysis showed substantial alignment between GPT-4o predictions and human judgments, suggesting that the annotation guidelines capture reasonably consistent persuasive and deceptive patterns within betting advertisements.

\begin{table}[h!]
\centering
\small
\renewcommand{\arraystretch}{1.2}
\begin{tabular}{lc}
\toprule
\textbf{Metric} & \textbf{Score} \\
\midrule
Cohen's $\kappa$ & 0.7421 \\
Macro-F1 & 0.7314 \\
Weighted-F1 & 0.7688 \\
Accuracy & 0.7763 \\
\bottomrule
\end{tabular}

\caption{Agreement between GPT-4o judgments and finalized human annotations}
\label{tab:llmjudge}
\end{table}

The lower agreement relative to human annotator consistency suggests that the LLM occasionally over-relied on surface-level sentiment and promotional tone when distinguishing between manipulative and deceptive content. In particular, advertisements with positive emotional framing but no explicit false claims were more difficult for the model to classify consistently.

\section{Bootstrap Confidence Intervals}
\label{app:confidence_intervals}

Table~\ref{tab:ci_full} reports 95\% bootstrap confidence intervals (10,000 resamples) for all models and metrics on the test set.

\begin{table*}[!t]
\centering
\footnotesize
\setlength{\tabcolsep}{4pt}
\begin{tabular}{llccc}
\toprule
\textbf{Model} & \textbf{Metric} & \textbf{Point Est.} & \textbf{95\% CI} & \textbf{Width} \\
\midrule
BERT & accuracy & 0.7989 & [0.7698, 0.8267] & 0.0569 \\
     & macro-F1 & 0.6889 & [0.6493, 0.7270] & 0.0777 \\
     & weighted-F1 & 0.8011 & [0.7717, 0.8298] & 0.0581 \\
     & micro-F1 & 0.7989 & [0.7698, 0.8280] & 0.0582 \\
\midrule
RoBERTa & accuracy & 0.7738 & [0.7434, 0.8029] & 0.0595 \\
        & macro-F1 & 0.6616 & [0.6234, 0.7001] & 0.0767 \\
        & weighted-F1 & 0.7787 & [0.7483, 0.8085] & 0.0602 \\
        & micro-F1 & 0.7738 & [0.7434, 0.8029] & 0.0595 \\
\midrule
DistilBERT & accuracy & 0.8148 & [0.7870, 0.8426] & 0.0556 \\
           & macro-F1 & 0.7078 & [0.6693, 0.7453] & 0.0760 \\
           & weighted-F1 & 0.8164 & [0.7878, 0.8441] & 0.0563 \\
           & micro-F1 & 0.8148 & [0.7870, 0.8426] & 0.0556 \\
\midrule
ELECTRA & accuracy & 0.8254 & [0.7976, 0.8519] & 0.0543 \\
        & macro-F1 & 0.7231 & [0.6822, 0.7610] & 0.0788 \\
        & weighted-F1 & 0.8262 & [0.7986, 0.8537] & 0.0551 \\
        & micro-F1 & 0.8254 & [0.7976, 0.8519] & 0.0543 \\
\midrule
ALBERT & accuracy & 0.8228 & [0.7950, 0.8492] & 0.0542 \\
       & macro-F1 & 0.7181 & [0.6767, 0.7569] & 0.0802 \\
       & weighted-F1 & 0.8214 & [0.7930, 0.8484] & 0.0554 \\
       & micro-F1 & 0.8228 & [0.7963, 0.8492] & 0.0529 \\
\midrule
Longformer & accuracy & 0.8386 & [0.8122, 0.8651] & 0.0529 \\
           & macro-F1 & 0.7157 & [0.6758, 0.7549] & 0.0791 \\
           & weighted-F1 & 0.8352 & [0.8069, 0.8618] & 0.0549 \\
           & micro-F1 & 0.8386 & [0.8108, 0.8651] & 0.0543 \\
\midrule
BART & accuracy & 0.7910 & [0.7606, 0.8188] & 0.0582 \\
     & macro-F1 & 0.6848 & [0.6454, 0.7223] & 0.0769 \\
     & weighted-F1 & 0.7941 & [0.7646, 0.8230] & 0.0584 \\
     & micro-F1 & 0.7910 & [0.7619, 0.8201] & 0.0582 \\
\midrule
T5 & accuracy & 0.8108 & [0.7817, 0.8386] & 0.0569 \\
   & macro-F1 & 0.6795 & [0.6405, 0.7186] & 0.0781 \\
   & weighted-F1 & 0.8087 & [0.7796, 0.8370] & 0.0574 \\
   & micro-F1 & 0.8108 & [0.7831, 0.8386] & 0.0555 \\
\midrule
GPT-4o (zero-shot) & accuracy & 0.8029 & [0.7738, 0.8307] & 0.0569 \\
                   & macro-F1 & 0.7054 & [0.6633, 0.7459] & 0.0826 \\
                   & weighted-F1 & 0.8020 & [0.7728, 0.8305] & 0.0577 \\
                   & micro-F1 & 0.8029 & [0.7738, 0.8307] & 0.0569 \\
\midrule
GPT-4o (few-shot) & accuracy & 0.8135 & [0.7857, 0.8400] & 0.0543 \\
                  & macro-F1 & 0.7143 & [0.6734, 0.7526] & 0.0792 \\
                  & weighted-F1 & 0.8138 & [0.7857, 0.8415] & 0.0558 \\
                  & micro-F1 & 0.8135 & [0.7857, 0.8413] & 0.0556 \\
\midrule
GPT-4o (CoT) & accuracy & 0.8069 & [0.7778, 0.8347] & 0.0569 \\
             & macro-F1 & 0.7040 & [0.6638, 0.7426] & 0.0788 \\
             & weighted-F1 & 0.8092 & [0.7805, 0.8375] & 0.0570 \\
             & micro-F1 & 0.8069 & [0.7791, 0.8347] & 0.0556 \\
\bottomrule
\end{tabular}
\caption{95\% bootstrap confidence intervals (10,000 resamples) for all models. CI widths for accuracy are $\approx$0.053-0.060, consistent with the test set size (756 samples). Macro-F1 CIs are wider ($\approx$0.077-0.083) due to higher variance on the minority deceptive class. Overlapping CIs between adjacent models are consistent with non-significant statistical tests.}
\label{tab:ci_full}
\end{table*}

\section{Procedure Involving Data Collection and Construction}
\label{app:procedure}

\begin{enumerate}

\item \textbf{Data Collection:}
Betting-related advertisements were collected from public social media platforms, including Instagram and Reddit, using platform search mechanisms and the Meta Ads Library. Relevant advertisements associated with betting applications and services were retrieved using betting-related keywords.

\item \textbf{Data Cleaning:}
Duplicate advertisements, irrelevant promotional content, spam entries, and non-betting advertisements were removed during preprocessing. Advertisements containing minimal or non-informative text were excluded from the dataset.

\item \textbf{Data Preprocessing:}
URLs, usernames, hashtags, excessive emojis, and formatting artifacts were normalized where necessary. Text preprocessing was performed while preserving the persuasive and emotional language patterns important for annotation.

\item \textbf{Manual Annotation:}
Each advertisement was independently reviewed and labeled in accordance with the annotation guidelines. Annotators also provided concise, human-written explanations of the reasoning behind each assigned label.

\item \textbf{Quality Verification:}
Annotations were reviewed for consistency, and disagreements were resolved through collaborative discussion before constructing the final dataset.

\item \textbf{Dataset Finalization:}
After quality verification, duplicate advertisements were removed and the final dataset was partitioned into train, validation, and test splits using stratified sampling.
\end{enumerate}

 \section{Longformer Ablation Study}
\label{app:longformer}

To justify Longformer's inclusion despite short median text length, we conducted an ablation study varying the maximum sequence length. Table~\ref{tab:longformer_ablation} shows that performance drops at 128 tokens due to truncation (5.2\% of texts affected), but increasing from 256 to 4096 tokens yields a marginal change. This confirms that the architectural benefit stems from Longformer's global attention mechanism rather than extended context length alone.

\begin{table*}[!t]
\centering
\small
\setlength{\tabcolsep}{3pt}
\renewcommand{\arraystretch}{1.1}
\resizebox{\textwidth}{!}{
\begin{tabular}{lcccp{4.5cm}}
\toprule
\textbf{max\_length} & \textbf{Acc} & \textbf{Ma-F1} & \textbf{W-F1} & \textbf{Notes} \\
\midrule
128  & 0.8254 & 0.6929 & 0.8247 & 5.2\% texts truncated \\
256  & 0.8386 & 0.7188 & 0.8332 & Paper setting (0.8\% truncated) \\
512  & 0.8373 & 0.7117 & 0.8335 & No additional benefit \\
4096 & 0.8373 & 0.7128 & 0.8326 & Full Longformer context \\
\bottomrule
\end{tabular}
}
\caption{Longformer performance across different maximum sequence lengths.}
\label{tab:longformer_ablation}
\end{table*}
% ─────────────────────────────────────────────────────────────
\section{Dataset Composition \& Risk Profile}
\label{app:dataset}
 
The dataset comprises 4,000 digital engagement entries collected from social media
platforms. Each entry was independently annotated and assigned a risk category based
on content type, language patterns, and psychological manipulation tactics.
 
\begin{table*}[t!]
\footnotesize
\centering
\caption{Dataset Risk Profile Summary}
\label{tab:dataset_composition}
\begin{tabular}{lrr}
\toprule
\textbf{Category} & \textbf{Count} & \textbf{Proportion} \\
\midrule
Safe / Low Risk (Sports \& Entertainment) & 2,038 & 50.9\% \\
Critical Risk (Gambling \& Betting)   & 1,962 & 49.1\% \\
\midrule
\textbf{Total}          & \textbf{4,000} & \textbf{100\%} \\
\bottomrule
\end{tabular}
\end{table*}
 
Sentiment analysis reveals a critical anomaly termed \textit{Deceptive Positivity}:
Gambling content maintains a near-identical positive sentiment score to benign sports
content that actively lowers users' cognitive defenses.
 
\begin{table*}[h!]
\centering
\caption{Mean Sentiment Scores by Content Category}
\label{tab:sentiment}
\begin{tabular}{lc}
\toprule
\textbf{Content Category} & \textbf{Mean Sentiment Score} \\
\midrule
Safe / Low Risk (Sports \& Entertainment) & $+0.15$ \\
Critical Risk (Gambling \& Betting)   & $+0.14$ \\
\bottomrule
\end{tabular}
\end{table*}
 
% ─────────────────────────────────────────────────────────────

\section{Sentiment and Emotional Tone}

\begin{figure*}[t]
  \centering
  \includegraphics[width=\linewidth]{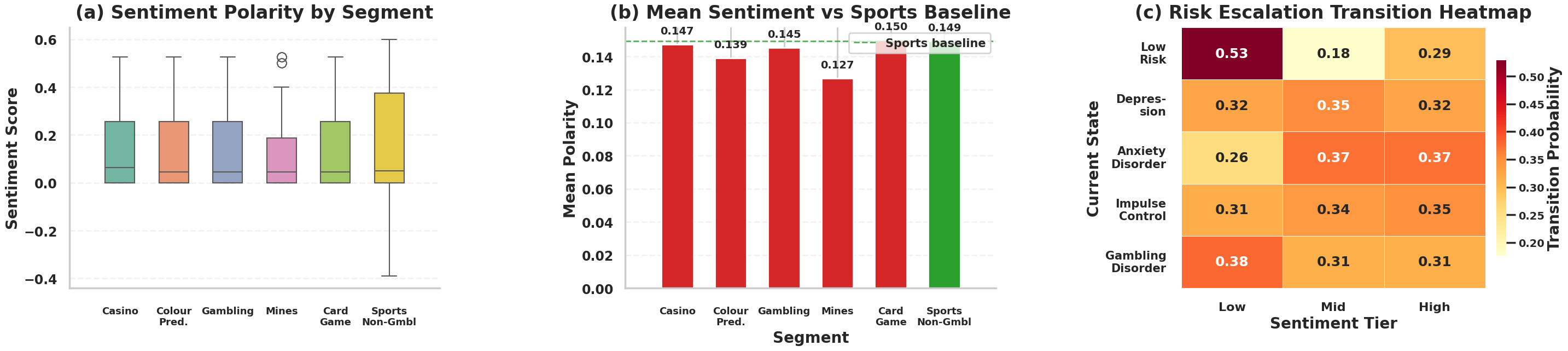}
  \caption{Sentiment, emotional tone, and the deceptive positivity effect: (a) Sentiment polarity by segment. (b) Mean 
  sentiment versus sports baseline. (c) Risk escalation transition 
  heatmap showing systematic progression toward gambling disorder 
  as sentiment becomes more positive.}
  \label{fig:rq2}
\end{figure*}

All five gambling segments maintain a narrow band of mild positivity (mean 0.127-0.150), nearly identical to benign sports content (Figure~\ref{fig:rq2}(a-b)). This is not neutral; it is calibrated. As the KDE overlap in Figure~\ref{fig:rq1}(c) confirms, gambling and sports content are effectively indistinguishable in the positive range, producing a Trojan Horse effect where harmful content mimics safe entertainment with no warning signal. This manufactured positivity explains GPT-4o's systematic difficulty with the \textit{deceptive} 
class: without fine-tuning, the model anchors to surface tone rather than the deeper rhetorical structure that distinguishes deception from 
manipulation.

\section{Neurological \& Psychological Mechanisms of Harm}
\label{app:mechanisms}
 
\subsection{Dopamine Dysregulation The Addiction Loop}
 
The gambling content in this dataset heavily employs \textit{Intermittent Variable
Rewards} (e.g., ``Try your luck,'' ``Big Wins Await''). Neurologically, the
uncertainty of the reward triggers a higher dopamine release than a guaranteed win,
conditioning users to seek the anticipatory ``high'' and rewiring neural pathways
analogous to substance abuse disorders.
 
\subsection{Prefrontal Cortex Shutdown:   Impulse Control Bypass}
 
Urgency triggers found throughout the dataset (``Hurry,'' ``Limited Time,''
``Register Now'') are designed to bypass the Prefrontal Cortex, the brain's center
for logic and long-term planning. This forces users into a \textit{Hot State},
precipitating impulsive financial decisions without regard for consequences.
 
\subsection{Illusion of Control: Cognitive Distortion}
 
Phrases such as ``Play Smart,'' ``Prediction,'' and ``Strategy'' are embedded in games of pure chance. This fosters a cognitive distortion in which users believe they can influence random outcomes, a primary driver of Ludomania (Compulsive Gambling Disorder), as losses are attributed to flawed strategy rather than mathematical inevitability.
 
% ─────────────────────────────────────────────────────────────
\section{Comprehensive Disorder Analysis}
\label{app:disorders}
 
Table~\ref{tab:disorders} categorizes the specific mental health disorders triggered
by the dataset content, their severity classification, primary linguistic triggers,
neurobiological mechanism, and clinical consequence.
 
\begin{table*}[ht]
\centering
\small
\renewcommand{\arraystretch}{1.3}

\caption{Mental Health Disorders Triggered by Gambling Promotion Content}
\label{tab:disorders}

\begin{tabularx}{\textwidth}{p{2.3cm} c p{2.3cm} X X}
\toprule
\textbf{Mental Health Disorder} &
\textbf{Risk Level} &
\textbf{Primary Triggers} &
\textbf{Mechanism of Action} &
\textbf{Clinical Consequence} \\
\midrule
 
Gambling Disorder (Ludomania) &
\textcolor{RiskSevere}{\textbf{SEVERE}} &
``Jackpot,'' ``Spin,'' ``Win,'' ``Luck'' &
Variable Ratio Reinforcement: hijacks the brain's dopamine reward system through uncertain outcomes. &
Pathological need to bet; inability to stop despite severe consequences. \\

Suicidal Ideation \& Risk &
\textcolor{RiskCritical}{\textbf{CRITICAL}} &
``Recover Losses,'' ``Deposit,'' high-stakes betting &
Serotonin/Dopamine Crash: chasing losses leads to sudden financial ruin and hopelessness. &
Sudden drop in mental stability; high risk of self-harm following major losses. \\
 
Impulse Control Disorder &
\textcolor{RiskHigh}{\textbf{HIGH}} &
``Hurry,'' ``Now,'' ``Limited,'' ``Exclusive'' &
Executive Function Bypass: urgency cues disable the logical Prefrontal Cortex. &
Inability to delay gratification; reckless behavior in finance and daily life. \\

Generalized Anxiety Disorder &
\textcolor{RiskHigh}{\textbf{HIGH}} &
``Deposit Match,'' ``Withdrawal,'' ``Bet'' &
Cortisol Dysregulation: volatility of wins/losses maintains a chronic Fight-or-Flight state. &
Panic attacks, sleep disturbances (insomnia), and chronic physical stress. \\
 
Depression (Situational) &
\textcolor{RiskHigh}{\textbf{HIGH}} &
Post-loss realization &
Reward System Burnout: the brain becomes desensitized to normal joys after high-dopamine gambling spikes. &
Anhedonia, lethargy, and persistent feelings of worthlessness. \\
 
Body Dysmorphia / Social Envy &
\textcolor{RiskModerate}{\textbf{MODERATE}} &
``Record-breaking,'' idealised athlete imagery &
Social Comparison Theory: constant exposure to idealized highlight reels of others. &
Low self-esteem, feelings of inadequacy, and a need for validation. \\
 
\bottomrule
\end{tabularx}
\end{table*}
 
% ─────────────────────────────────────────────────────────────
\section{Criticality Assessment: Suicidal Risk Factor}
\label{app:suicide}
 
The correlation between gambling promotion content and suicide risk renders this
The dataset highlights critically high-risk behavioral patterns. The mechanism proceeds through three identifiable
stages:
 
\begin{enumerate}
  \item \textbf{The Chase Phase.} Content encourages users to ``recover'' losses by
    making further deposits, sustaining engagement through loss-aversion
    psychology.
 
  \item \textbf{The Crash.} When mathematically inevitable losses occur, the user
    experiences a rapid depletion of both dopamine and serotonin.
 
  \item \textbf{The Danger Zone.} This neurochemical crash, combined with the
    tangible reality of debt or bankruptcy, creates a window of acute
    suicidality. Research indicates that individuals with Gambling Disorder have
    the highest suicide attempt rates among all addiction demographics, with
    estimates reaching up to 20\%~\cite{blaszczynski2002pathways}.
\end{enumerate}
 
% ─────────────────────────────────────────────────────────────
\section{Visual Evidence \& Analytical Interpretation}
\label{app:visuals}
 
Five core visual artifacts were generated to support the clinical and psychological
risk assessment. Each figure is interpreted in the context of the dataset.
 
\subsection{Integrated Risk \& Severity Analysis}
 
A multi-panel composite figure consolidating four complementary analytical views:
Risk Severity Distribution by Disorder; Sentiment vs.\ Severity Density Mapping;
Risk Escalation Transition Heatmap and Cumulative Contribution to Overall Mental
Health Risk. Gambling Disorder and Suicidal Ideation exhibit the highest median
severity with the widest dispersion, indicating instability and escalation risk.
Transition probabilities confirm systematic movement from Low Risk $\rightarrow$
Depression $\rightarrow$ Gambling Disorder.
 
\subsection{Psychological Attack Vector Composition}
 
A stacked bar chart breaking down psychological triggers across gambling and
non-gambling content. Gambling content relies heavily on financial coercion (55.8\%),
followed by urgency and cognitive distortion. Non-gambling content lacks financial
manipulation entirely, confirming that gambling content is structurally engineered as
a psychological attack.
 
\subsection{Sentiment Overlap: The Deceptive Mask}
 
A KDE plot comparing sentiment distributions of critical gambling content and
low-risk sports content. Strong overlap exists in the positive sentiment range
($0.0$-$0.3$), visually confirming the \textit{Trojan Horse Effect}: harmful content
intentionally mimics the tone of sports and entertainment, rendering sentiment-based
detection insufficient as a sole classifier.
 
\subsection{Psychological Impact Comparison Heatmap}
 
A heatmap comparing mental health disorders across five psychological impact
dimensions. Gambling Disorder and Suicidal Ideation show extreme reward sensitivity
dysregulation. Decision stability collapses most sharply in gambling-linked disorders.
Reward system hijacking emerges as the primary neuropsychological mechanism driving
severe outcomes.
 
\subsection{Cognitive \& Neurobehavioural Radar Profile}
 
A radar chart visualizing cognitive dysfunction patterns across disorders. Gambling
Disorder and Suicidal Ideation produce spiked, asymmetric profiles indicating
instability, whereas anxiety and depression show more balanced but persistent
impairment. Decision stability is most compromised in gambling-related conditions,
highlighting a qualitatively distinct harm profile.
 
% ─────────────────────────────────────────────────────────────
\section{Summary of Clinical Findings}
\label{app:summary}
 
All five visual artifacts and the quantitative disorder analysis converge on a single conclusion: gambling content in this dataset operates as a high-risk psychological system, not neutral entertainment, and significantly elevates the probability of severe mental health outcomes, including suicidal ideation.
 
The dataset demonstrates a \textit{Dual-Diagnosis Trigger} structure, simultaneously
promoting social inadequacy via sports comparison and severe addiction via gambling predation. The juxtaposition of safe entertainment with high-risk financial triggers the Trojan Horse Effect, normalizes gambling behavior, and lowers user psychological defenses, exploiting neurological reward mechanisms without adequate
warning or consent.
 
\end{document}